\documentclass[10pt,twocolumn,letterpaper]{article}

\usepackage[pagenumbers]{cvpr}   

\usepackage{booktabs}
\usepackage{tabularx}
\usepackage{array}
\usepackage{amsmath}

\definecolor{cvprblue}{rgb}{0.21,0.49,0.74}
\usepackage[pagebackref,breaklinks,colorlinks,allcolors=cvprblue]{hyperref}

\def\paperID{*****}
\def\confName{CVPR}
\def\confYear{2026}

\newcolumntype{L}[1]{>{\raggedright\arraybackslash}p{#1}}
\newcolumntype{Y}{>{\raggedright\arraybackslash}X}

\makeatletter
\newcommand{\blfootnote}[1]{%
    \begingroup
    \refstepcounter{footnote}%
    \renewcommand\@makefnmark{}%
    \footnotetext[\value{footnote}]{#1}%
    \addtocounter{footnote}{-1}%
    \endgroup
}
\makeatother

\title{Training-Free Model Ensemble for Single-Image Super-Resolution via Strong-Branch Compensation}

\author{
Gengjia Chang$^{1}$ \quad Xining Ge$^{2}$ \quad Weijun Yuan$^{3}$ \quad Zhan Li$^{3}$\\
Qiurong Song$^{4}$ \quad Luen Zhu$^{4}$ \quad Shuhong Liu$^{5,\dagger}$\\
$^{1}$Hefei University of Technology \quad
$^{2}$Hangzhou Dianzi University \quad
$^{3}$Jinan University\\
$^{4}$South China Agricultural University \quad
$^{5}$The University of Tokyo
}

\begin{document}
\maketitle
\blfootnote{The \emph{NTIRE 2026 Image Super-Resolution (x4)} challenge is hosted on Codabench. Challenge page: \url{https://www.codabench.org/competitions/13516/}.}

\begin{abstract}
Single-image super-resolution has progressed from deep convolutional baselines to stronger Transformer and state-space architectures, yet the corresponding performance gains typically come with higher training cost, longer engineering iteration, and heavier deployment burden. In many practical settings, multiple pretrained models with partially complementary behaviors are already available, and the binding constraint is no longer architectural capacity but how effectively their outputs can be combined without additional training. Rather than pursuing further architectural redesign, this paper proposes a training-free output-level ensemble framework. A dual-branch pipeline is constructed in which a Hybrid attention network(HAT) with TLC inference provides stable main reconstruction, while a MambaIRv2 branch with geometric self-ensemble supplies strong compensation for high-frequency detail recovery. The two branches process the same low-resolution input independently and are fused in the image space via a lightweight weighted combination, without updating any model parameters or introducing an additional trainable module. As our solution to the NTIRE 2026 Image Super-Resolution ($\times 4$) Challenge, the proposed design consistently improves over the base branch and slightly exceeds the strong branch alone in PSNR at the best operating point under a unified DIV2K bicubic $\times 4$ evaluation protocol. Ablation studies confirm that output-level compensation provides a low-overhead and practically accessible upgrade path for existing super-resolution systems.
\end{abstract}

\section{Introduction}
Image super-resolution is a fundamental task in low-level vision that underpins a wide range of downstream applications \cite{ren2026esr}, including autonomous driving~\cite{lisgs2025,liumg2025,zhou2024mod}, VR/AR~\cite{lidense2025}, 3D reconstruction~\cite{liu2025i2nerf,cui2026unifying,liu20263drr}, and scene understanding given low-quality images~\cite{liuderain2025,liu2025realx3d,liudenoise2026,ge2026clip}. In adverse-condition 3D pipelines, this demand becomes even more explicit: smoke and extreme low-light severely affect cross-view consistency, geometry recovery, and smoke-free novel-view synthesis~\cite{zheng20263d,liu2026elog,fu2026smokegs,cao2026gensmoke,zhu2026naka,chen2026dehaze,guo2026reliability}. Single-image super-resolution (SISR) aims to recover a high-resolution image from a low-resolution observation and remains a representative problem in this domain. Transformer-based restorers such as SwinIR and IPT established strong early baselines~\cite{liang2021swinir,chen2021pre}, and later models including Uformer, Restormer, and NAFNet improved the balance between quality and efficiency~\cite{wang2022uformer,zamir2022restormer,chen2022simple}. SR-specific designs such as HAT, ELAN, DAT, and SRFormer further strengthened attention and aggregation~\cite{conde2022swin2sr,chen2023activating,zhang2022efficient,chen2023dual,zhou2023srformer}, while state-space models MambaIR and MambaIRv2 introduced yet another strong design line~\cite{guo2024mambair,guo2025mambairv2}. These advances have clearly improved reconstruction fidelity, but they also raise a practical question: once a capable SR system is already in place, is retraining a stronger model always the most efficient path to further gains?

In many practical settings, the answer is no. Multiple pretrained models with different inductive biases are often already available, and the binding constraint is no longer architectural capacity alone, but how effectively their complementary strengths can be combined without additional training. The broader trend toward lightweight and deployment-aware SR reflects the same concern~\cite{wang2021exploring,li2022blueprint,sun2023spatially,choi2023n,wang2023omni,zhu2021lightweight,fang2022hybrid,gao2022lightweight,park2021dynamic,gao2022feature,feng2022lkasr,li2023dlgsanet,li2023cross,zhang2024hit,li2025srconvnet}. Real-world and blind SR further reinforces this view, where practical degradation handling and post hoc correction matter as much as raw backbone capacity~\cite{wang2021real,chen2022real,wei2021unsupervised,liang2022details,liang2022efficient,wu2024seesr,liu2022blind,luo2022deep,yue2022blind,zhang2021designing,zhou2023learning,liu2024degradation,yang2024dynamic,liu2024cdformer}. Recent diffusion and arbitrary-scale SR work similarly suggests that stronger output quality can arise from more capable inference procedures rather than from an entirely new backbone~\cite{li2022srdiff,wang2024exploiting,shang2024resdiff,wang2024sinsr,cao2023ciaosr,yao2023local,song2023ope,wang2023deep,gendy2023lightweight,sun2024ensir}. Related evidence in adjacent restoration tasks points in the same direction: dual-branch complementary modeling has recently been explored for remote-sensing infrared SR, while data-centric training and self-ensemble remain effective levers in denoising and night-time dehazing~\cite{ge2026dual,chang2026beyond,ge2026clip}.

Motivated by this observation, this paper proposes a training-free dual-branch output-level ensemble for SISR. The two branches are treated asymmetrically: a Hybrid attention network with TLC inference serves as the base reconstruction path, while a MambaIRv2 branch with geometric self-ensemble provides strong compensation for high-frequency detail recovery. Compensation is performed directly in the output image space via a lightweight weighted combination, without updating any model parameters or introducing an additional trainable module.

The main contributions are as follows. First, we present a training-free dual-branch output-level compensation framework that upgrades an existing SR system without retraining any parameters. Second, we explicitly formulate the strong branch as a compensation source rather than a symmetric peer, improving both interpretability and deployment clarity. Third, under a unified DIV2K bicubic $\times 4$ evaluation on 200 images, quantitative comparison, weight sweeping, and visual analysis confirm that even a small proportion of strong-branch injection yields measurable gains.

We also report our results in the NTIRE 2026 Image Super-Resolution ($\times 4$) Challenge \cite{ntire26srx4}, where our team \textbf{wedream} ranked \textbf{2nd}, achieving 33.47 dB PSNR and 0.9105 SSIM, providing independent external validation of the effectiveness of the proposed training-free ensemble strategy.

\begin{figure*}[!t]
    \centering
    \includegraphics[width=0.92\textwidth]{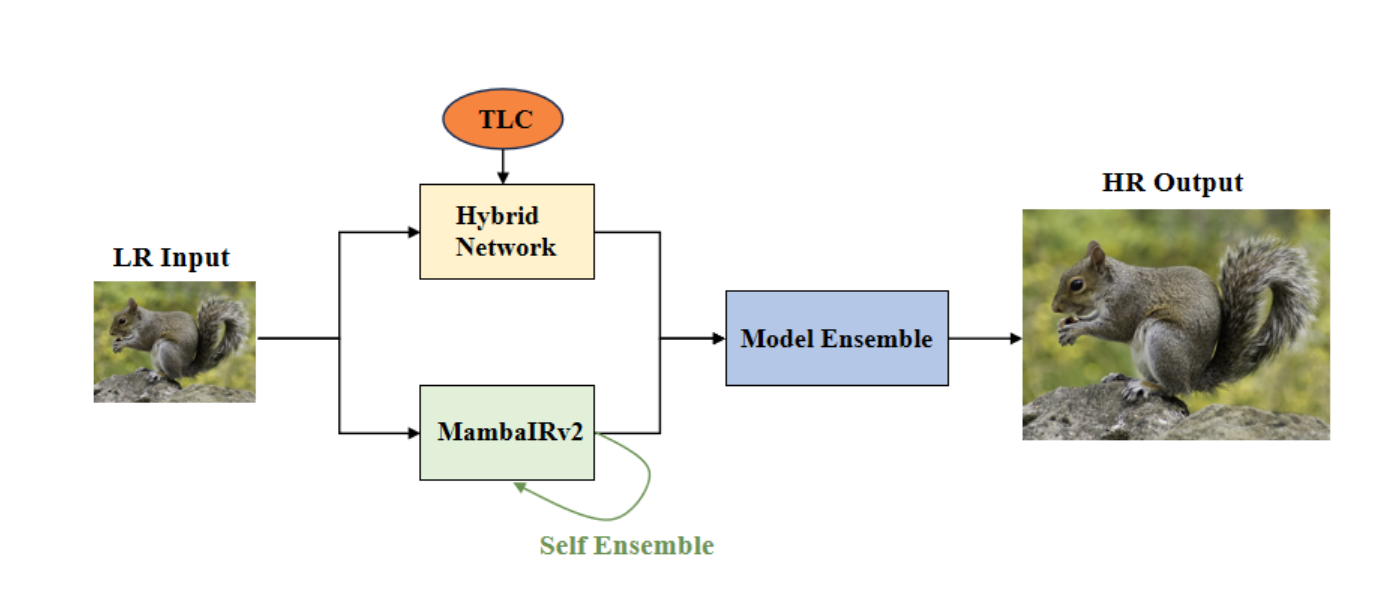}
    \caption{Overview of the training-free output-level ensemble. The HAT + TLC branch provides the main reconstruction, while the MambaIRv2 branch contributes stronger detail recovery through self-ensemble. The final result is obtained by weighted image-space compensation.}
    \label{fig:pipeline}
\end{figure*}

\section{Related Work}
\subsection{High-Performance SISR Backbones}
Broad surveys note that SISR progress is driven not only by larger backbones, but also by changing assumptions about efficiency and deployment~\cite{bashir2021comprehensive,gendy2023lightweight}. Among transformer-era methods, SwinIR and IPT established strong early baselines~\cite{liang2021swinir,chen2021pre}. Uformer, Restormer, and NAFNet then broadened the restoration design space with stronger long-range modeling and simpler optimization behavior~\cite{wang2022uformer,zamir2022restormer,chen2022simple}. Other SR-oriented models explored transformer formulations more directly, including Transformer for SISR and non-local sparse attention designs~\cite{lu2022transformer,mei2021image}. 

Subsequent work focused on more specialized attention and aggregation designs. Swin2SR and HAT are representative examples of this trend~\cite{conde2022swin2sr,chen2023activating}. ELAN, DAT, and SRFormer further show that local-global interaction can be improved through more structured attention design~\cite{zhang2022efficient,chen2023dual,zhou2023srformer}. In parallel, MambaIR and VMamba suggest that state-space modeling can serve as a viable alternative to pure Transformer scaling~\cite{guo2024mambair,liu2024vmamba}. MambaIRv2 strengthens this restoration-oriented state-space line even further~\cite{guo2025mambairv2}. These works collectively suggest that no single architectural family dominates every quality-efficiency tradeoff, which is precisely why combining pretrained branches remains attractive.

Efficient SR offers another relevant context. Sparse inference and compact blueprint designs are clear examples~\cite{wang2021exploring,li2022blueprint}. Adaptive modulation, N-Gram Swin variants, and omni aggregation continue the same lightweight direction~\cite{sun2023spatially,choi2023n,wang2023omni}. Hybrid CNN-transformer designs and dynamic lightweight attention provide another branch of efficient SR research~\cite{zhu2021lightweight,fang2022hybrid,gao2022lightweight,park2021dynamic}. Feature-distillation weighting, large-kernel attention, local-global self-attention, focused inference, and hierarchical efficient transformers further enrich this line~\cite{gao2022feature,feng2022lkasr,li2023dlgsanet,li2023cross,zhang2024hit,li2025srconvnet}. All of these works reinforce the idea that SR systems are increasingly judged by upgrade cost as well as by pure reconstruction quality.

\subsection{Inference-Time Enhancement and Model Combination}
Inference-time enhancement is attractive because it improves restoration quality without retraining. TLC is a representative example: it reduces train-test inconsistency by converting global aggregation into local operations at test time~\cite{chu2022improving}. Geometric self-ensemble is another practical enhancement widely used when robustness matters. Such methods improve a single pretrained model, but they do not directly explain how multiple heterogeneous pretrained branches should be combined when retraining is undesirable. Adjacent restoration settings provide a useful reference point here: dual-branch remote-sensing infrared SR demonstrates the value of complementary branches for balancing structure recovery and detail enhancement, while recent denoising and night-time dehazing pipelines further show that data-centric optimization and self-ensemble can release extra performance from a mature backbone~\cite{ge2026dual,chang2026beyond,ge2026clip}.

This question becomes particularly relevant in real-world and blind SR. Real-ESRGAN, domain-aware real-world SR, and related reviews all highlight the gap between benchmark SR and practical degradation handling~\cite{wang2021real,chen2022real,wei2021unsupervised}. Artifact-detail tradeoff and degradation-adaptive restoration further show that post hoc correction can be as important as raw backbone strength~\cite{liang2022details,liang2022efficient,wu2024seesr}. Blind SR also studies constrained estimation and explicit degradation modeling~\cite{liu2022blind,luo2022deep,yue2022blind}. Practical degradation design, correction filters, degradation-aware transformers, and later kernel-prior or diffusion-style blind SR continue the same direction~\cite{zhang2021designing,zhou2023learning,liu2024degradation,yang2024dynamic,liu2024cdformer}.

Diffusion and arbitrary-scale SR provide another useful perspective. Diffusion-based SR shows that stronger results can emerge from better iterative generation and restoration priors~\cite{li2022srdiff,wang2024exploiting,shang2024resdiff,wang2024sinsr}. Arbitrary-scale and implicit SR show a similar shift toward representation design through attention-in-attention, local flows, orthogonal position encoding, and scale-equivariant reconstruction~\cite{cao2023ciaosr,yao2023local,song2023ope,wang2023deep}. More generally, EnsIR demonstrates that training-free aggregation is already effective for image restoration~\cite{sun2024ensir}. Unlike symmetric model averaging, our work focuses on output-level compensation around a designated base branch: the strong branch is introduced to correct the base result instead of replacing it or participating as an equal branch in a learned fusion module. In that sense, the method sits at the intersection of inference-time enhancement, model reuse, and deployment-oriented SR upgrading.

\section{Method}
\subsection{Problem Formation}
Given a low-resolution input image $I_{\mathrm{LR}}$, our goal is to produce a high-quality super-resolved image $I_{\mathrm{SR}}$ without additional training. We assume that two pretrained branches are already available and exhibit complementary behavior. The first is a base reconstruction branch that is stable on the main structure. The second is a stronger branch that is more capable of recovering difficult details. The central design question is how to inject the benefit of the stronger branch while keeping the whole system training-free.

\subsection{Overall Framework}
Figure~\ref{fig:pipeline} illustrates the proposed framework. The base branch uses HAT + TLC to generate the main reconstruction:
\begin{equation}
I_{\mathrm{base}} = F_{\mathrm{b}}^{\mathrm{TLC}}(I_{\mathrm{LR}}).
\end{equation}
The strong branch uses MambaIRv2 with $K=8$ geometric self-ensemble:
\begin{equation}
I_{\mathrm{strong}} = \frac{1}{K}\sum_{k=1}^{K} T_k^{-1}\big(F_{\mathrm{s}}(T_k(I_{\mathrm{LR}}))\big).
\end{equation}
The final output is computed by a lightweight image-space combination
\begin{equation}
I_{\mathrm{SR}} = (1-\alpha) I_{\mathrm{base}} + \alpha I_{\mathrm{strong}},
\end{equation}
where $\alpha \in [0,1]$ is the weight assigned to the strong branch.

This design is intentionally asymmetric. The base branch provides a reliable starting point, while the strong branch behaves as a compensation source for harder details. Compared with feature-level interaction or learned fusion, this output-space formulation has two advantages. First, the deployment path is simple because the original pretrained branches remain unchanged. Second, the contribution of the strong branch is directly observable through the scalar weight $\alpha$, which makes the method easy to analyze.

\subsection{Base Reconstruction Branch}
The base branch is summarized as HAT + TLC. Methodologically, its role is not to chase the highest possible detail sharpness, but to provide stable structure recovery and a reliable starting prediction. TLC is applied only at inference time, so the base branch can better handle local statistical variation without changing model parameters. In our formulation, the base branch should be understood as the anchor of the final prediction rather than as a weak baseline to be replaced completely.

\subsection{Strong Compensation Branch}
The strong branch uses MambaIRv2 combined with self-ensemble. Its role is to provide more aggressive high-frequency modeling and stronger local detail priors than the base branch. Because self-ensemble aggregates multiple transformed predictions, the strong branch becomes more stable and thus more suitable as a compensation source. Importantly, all operations still remain in the inference stage; no additional optimization or branch-specific fine-tuning is introduced.

\subsection{Why Output-Level Compensation}
We deliberately avoid learning an extra fusion head. In many application scenarios, users already have deployed SR models and simply want a low-modification way to improve the current system. Output-level compensation fits this requirement because it does not alter pretrained weights, does not require new training data, and keeps the interpretation straightforward: the final image is a weighted combination of a stable base estimate and a stronger detail-oriented estimate. The price is extra inference cost due to the dual-branch design, so the method is best viewed as a quality-oriented upgrade path rather than a speed-oriented replacement for single-branch inference.

\subsection{Compensation Weight Analysis}
\begin{center}
    \includegraphics[width=\columnwidth]{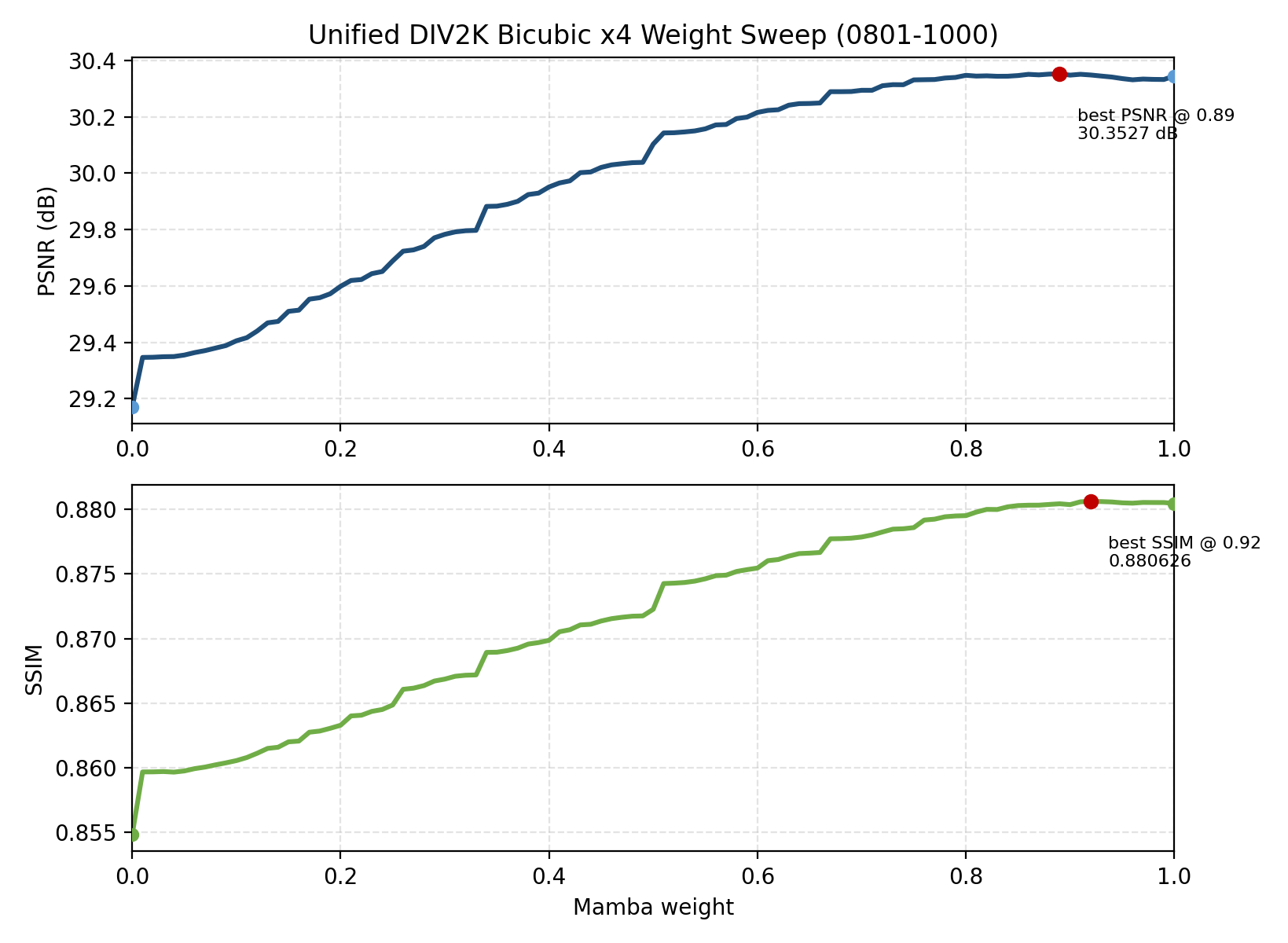}
    \captionsetup{hypcap=false}
    \captionof{figure}{PSNR and SSIM curves under different strong-branch weights.}
    \label{fig:weights}
\end{center}

\begin{figure*}[!t]
    \centering
    \includegraphics[width=0.94\textwidth]{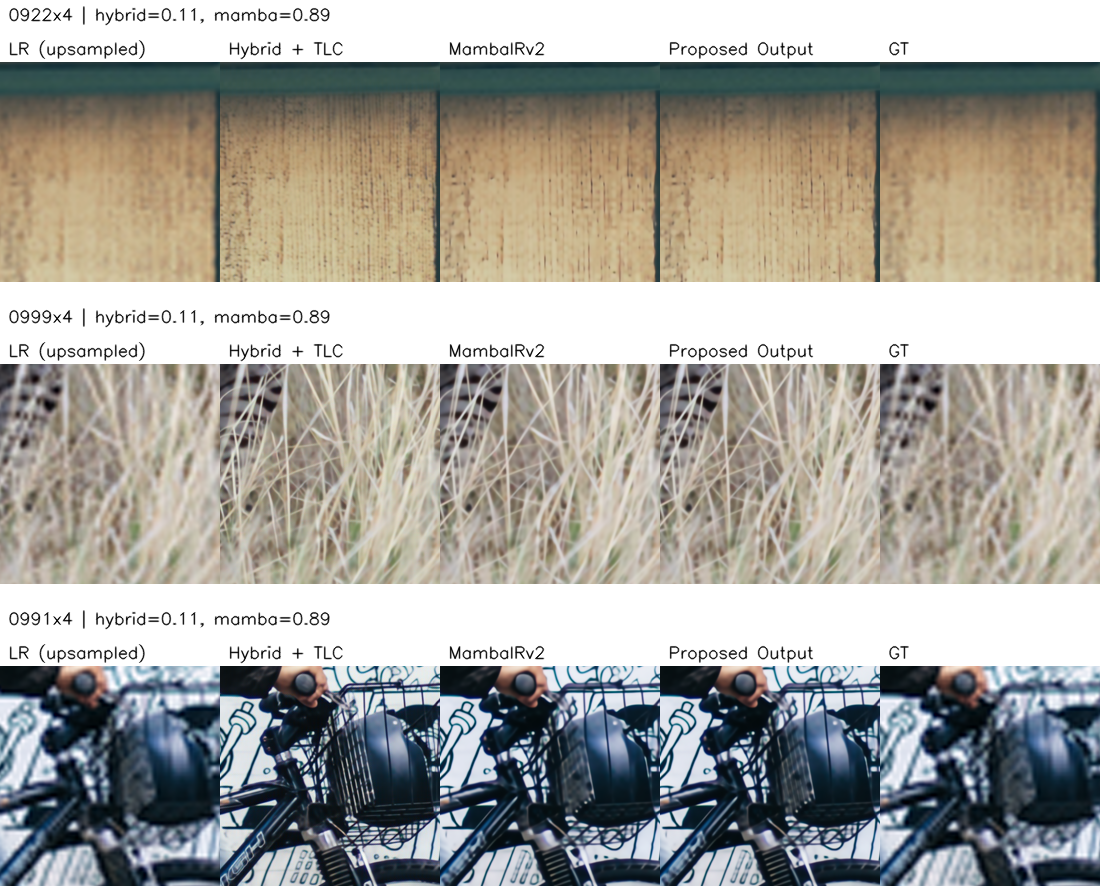}
    \caption{Qualitative comparison on representative DIV2K $\times 4$ samples. From left to right: LR input, HAT + TLC, MambaIRv2, proposed output, and ground truth.}
    \label{fig:qualitative}
\end{figure*}

\section{Experiments}
\subsection{Experimental Setting}
Experiments are conducted under a unified DIV2K bicubic $\times 4$ evaluation using 200 images indexed from 0801 to 1000. Low-resolution inputs are generated by bicubic downsampling and compared against the corresponding high-resolution targets. We report PSNR and SSIM as the main quantitative metrics. The comparison includes three representative reconstructions: the base HAT + TLC branch, the strong MambaIRv2 + self-ensemble branch, and the proposed compensated result.

No retraining, extra fine-tuning, or learnable fusion module is introduced. The main quantitative result uses the compensation weight that achieves the best PSNR under the current evaluation protocol. We further analyze the continuous performance change as the strong-branch weight varies from 0 to 1.

\subsection{Main Quantitative Results}
\begin{table}[h]
    \centering
    \small
    \setlength{\tabcolsep}{4pt}
    \renewcommand{\arraystretch}{1.10}
    \caption{Main results under the unified DIV2K bicubic $\times 4$ setting.}
    \label{tab:main-results}
    \begin{tabular}{lcc}
        \toprule
        Method & PSNR & SSIM \\
        \midrule
        HAT + TLC & 29.1696 & 0.854802 \\
        MambaIRv2 + self-ensemble & 30.3451 & 0.880466 \\
        Ours (0.11 / 0.89) & \textbf{30.3527} & 0.880438 \\
        \bottomrule
    \end{tabular}
\end{table}

Table~\ref{tab:main-results} reports the main comparison. The proposed output-level compensation substantially improves the base branch and also slightly surpasses the strong branch alone in PSNR at the best operating point. Specifically, the base branch achieves 29.1696 dB / 0.854802, while the strong branch reaches 30.3451 dB / 0.880466. The compensated result at Hybrid $=0.11$ and Mamba $=0.89$ further reaches 30.3527 dB / 0.880438.

Compared with the base branch, the compensated result improves PSNR by 1.1831 dB and SSIM by 0.025636. Compared with the strong branch, the gain is smaller but still meaningful in PSNR, which indicates that the base branch retains complementary information that is not fully covered by the stronger model alone. This behavior supports the central argument of the paper: heterogeneous pretrained SR branches can provide useful complementarity even when no extra fusion learning is allowed.

Figure~\ref{fig:weights} shows the PSNR and SSIM curves when the Mamba branch weight varies continuously. Two observations are important. First, even very small strong-branch contribution brings measurable improvement over the pure base configuration. This means the strong branch is not only useful in the near-strong regime, but can act as a stable compensation source over a broad interval. Second, the best PSNR appears in the high-weight regime of the MambaIRv2 branch at 0.89, while SSIM remains competitive and reaches its own best region around 0.92. Therefore, the method does not require a fragile narrow optimum to work well.

\subsection{Qualitative Results}
Figure~\ref{fig:qualitative} presents representative visual comparisons. In samples such as 0922$\times 4$, 0999$\times 4$, and 0991$\times 4$, the compensated result preserves the global structure of the base branch while absorbing the stronger local texture recovery of MambaIRv2. The gain is especially visible on repeated patterns, local edges, and texture continuity. These examples show that the compensation branch does not merely sharpen the base result indiscriminately; instead, it improves local detail quality while keeping the main structure stable.

\subsection{Discussion and Limitations}
The method is attractive because it is practical. When two pretrained branches already exist, output-level compensation yields a low-overhead upgrade path with no additional training, no new fusion parameters, and a transparent control variable. At the same time, some limitations should be stated clearly. The method increases inference latency because both branches must run at test time. The current evaluation is carried out under a unified DIV2K bicubic $\times 4$ setting, so conclusions should be interpreted within that scope. Finally, the present design uses a fixed scalar weight; more adaptive spatial or content-aware compensation could be interesting in future work, but would also reduce the simplicity that makes the current framework useful.

\section{Conclusion}
This paper presents our solution to the NTIRE 2026 Image Super-Resolution ($\times 4$) Challenge, proposing a training-free output-level ensemble framework for single-image super-resolution. By combining HAT as the base branch and a MambaIRv2 branch with geometric self-ensemble as the strong compensation path, the method improves a capable SR system without retraining any model parameters or learning an additional fusion module. Experiments on a unified DIV2K bicubic $\times 4$ protocol show that the compensated result clearly improves over the base branch and slightly outperforms the strong branch alone in PSNR at the best operating point. The results confirm that output-level compensation is a practical and deployment-friendly upgrade path for existing super-resolution systems.

{\small
\bibliographystyle{plain}
\bibliography{references}
}

\end{document}